\documentclass[letterpaper,10pt,conference]{ieeeconf}
\IEEEoverridecommandlockouts
\usepackage[utf8]{inputenc}

\usepackage[british]{babel}

\usepackage{amssymb}
\usepackage{amsmath}
\usepackage{array}

\usepackage{hyperref}
\usepackage{cleveref}
\usepackage{graphicx}
\usepackage{latexsym}
\usepackage{algorithm}
\usepackage{algpseudocode}
\usepackage{multicol}
\usepackage{tabularx}
\usepackage{booktabs}
\let\labelindent\relax
\usepackage{enumitem}
\usepackage{listings}
\usepackage{subcaption}
\usepackage{xcolor}
\usepackage[binary-units,detect-weight,detect-family,range-phrase=/]{siunitx}
\usepackage[export]{adjustbox} 
\usepackage{cite} 
\usepackage{paralist} 
\usepackage[T1]{fontenc} 
\usepackage{xspace} 

\usepackage{url}

\usepackage[acronym]{glossaries}
\glsdisablehyper 

\newacronym{cpg}{CPG}{Central Pattern Generator}
\newacronym{mimo}{MIMO}{Multiple-Input Multiple-Output}
\newacronym{nn}{NN}{Neural Network}
\newacronym{mlp}{MLP}{Multilayer Perceptron}
\newacronym{rnn}{RNN}{Recurrent Neural Network}
\newacronym{to}{TO}{Trajectory Optimisation}
\newacronym{lip}{LIP}{Linear Inverted Pendulum}
\newacronym{com}{CoM}{Center of Mass}
\newacronym{cop}{CoP}{Centre of Pressure}
\newacronym{zmp}{ZMP}{Zero-Moment Point}
\newacronym{pid}{PID}{Proportional-Integrative-Derivative}
\newacronym{rl}{RL}{Reinforcement Learning}
\newacronym{sea}{SEA}{Series-Elastic Actuator}
\newacronym{ga}{GA}{Genetic Algorithms}
\newacronym{pso}{PSO}{Particle Swarm Optimisation}
\newacronym{dof}{DOF}{Degree of Freedom}
\newacronym{mlr}{MLR}{Mesencephalic Locomotor Region}
\newacronym{haa}{HAA}{Hip Adduction/Abduction}
\newacronym{hfe}{HFE}{Hip Flexion/Extension}
\newacronym{kfe}{KFE}{Knee Flexion/Extension}
\newacronym{cma}{CMA-ES}{Covariance-Matrix Adaptation Evolution Strategy}
\newacronym{nsga}{NSGA-II}{Non-dominated Sorting Genetic Algorithm II}
\newacronym{es}{ES}{Evolution Strategy}
\newacronym{ros}{ROS}{Robotic Operating System}
\newacronym{cnn}{CNN}{Convolutional Neural Network}
\newacronym{ppo}{PPO}{Proximal Policy Optimisation}
\newacronym{trpo}{TRPO}{Trust Region Policy Optimisation}
\newacronym{bptt}{BPTT}{Backpropagation Through Time}
\newacronym{dl}{DL}{Deep Learning}
\newacronym{snn}{SNN}{Spiking Neural Network}

\hypersetup{
    pdfauthor   = {Luigi Campanaro},%
    pdftitle    = {CPG-Actor: Reinforcement Learning for Central Pattern Generators},%
    pdfsubject  = {Robotics},%
    pdfkeywords = {Central Pattern Generators, Reinforcement Learning, Legged Robots}%
}

\crefname{table}{Tab.}{Tab.s}
\crefname{figure}{Fig.}{Fig.s}
\Crefname{figure}{Figure}{Figures}
\crefname{section}{Sec.}{Sec.s}
\Crefname{section}{Section}{Sections}

\newcommand{\cpgactor}{\textsc{CPG-Actor}\xspace}

\title{\LARGE \bf
    \cpgactor: Reinforcement Learning for Central Pattern Generators
}
\author{
    Luigi Campanaro, Siddhant Gangapurwala, Daniele De Martini, Wolfgang Merkt and Ioannis Havoutis
    \thanks{All authors are with the Oxford Robotics Institute, University of Oxford, UK. Emails: \texttt{\{luigi, siddhant, daniele, wolfgang, ioannis\}@robots.ox.ac.uk}.}%
}

\begin{document}
\bstctlcite{IEEEexample:BSTcontrol}

\maketitle

\begin{abstract}
\glspl{cpg} have several properties desirable for locomotion: they generate smooth trajectories, are robust to perturbations and are simple to implement.
Although conceptually promising, we argue that the full potential of \glspl{cpg} has so far been limited by insufficient sensory-feedback information.
This paper proposes a new methodology that allows tuning \gls{cpg} controllers through gradient-based optimisation in a \gls{rl} setting. 
To the best of our knowledge, this is the first time \glspl{cpg} have been trained in conjunction with a \gls{mlp} network in a Deep-\gls{rl} context.
In particular, we show how \glspl{cpg} can directly be integrated as the Actor in an Actor-Critic formulation. 
Additionally, we demonstrate how this change permits us to integrate highly non-linear feedback directly from sensory perception to reshape the oscillators' dynamics. 
Our results on a locomotion task using a single-leg hopper demonstrate that explicitly using the \gls{cpg} as the Actor rather than as part of the environment results in a significant increase in the reward gained over time (\num{6}x more) compared with previous approaches.
Furthermore, we show that our method without feedback reproduces results similar to prior work with feedback.
Finally, we demonstrate how our closed-loop \gls{cpg} progressively improves the hopping behaviour for longer training epochs 
relying only on basic reward functions.
\end{abstract}


\glsresetall


\section{Introduction} \label{sec:introduction}

The increased manoeuvrability associated with legged robots in comparison to wheeled or crawling robots necessitates complex planning and control solutions.
Particularly, the requirement to maintain balance while interacting with an uncertain environment under noisy sensing severely restricts the time algorithms can spend on computing new solutions in response to perturbation or changes in the environment.
This greater complexity is further increased due to the high dimensionality of the problem, uncertainty about the environment, robot models and physical constraints. 
The current state-of-the-art for high-performance locomotion are modular, model-based controllers which break down the control problem in different sub-modules~\cite{kalakrishnan2011,bellicoso2018}: first, trajectory optimisation defines a motion plan over a longer time horizon using approximated models for computational efficiency; this plan is then tracked using advanced whole-body controllers which operate using the full dynamics model and provide robustness to external disturbances.
This rigorous approach is rooted in the knowledge of every portion of the motion, but it is also limited by heuristics handcrafted by engineers at each of the stages.
In fact, many systems need to estimate the ground contact or the slippage to trigger the transition between states or reflexes \cite{Camurri2017,Focchi2018}.
Such estimation is often based on heuristically-set thresholds, yet it is sensitive to unmodelled aspects of the environment.

Often the computations behind these controllers are so expensive that dealing with sudden disturbances is beyond their abilities and simplifications of the dynamic models are needed to meet the re-planning time requirements, resulting in a loss of dynamism and performances \cite{Ponton2018}. 

While the field of legged robot control has been dominated over the last decades by conventional control approaches, recently, data-driven methods demonstrated unprecedented results that outpaced most of the classical approaches in terms of robustness and dynamic behaviours \cite{Hwangbo2019,Joonho2020,gangapurwala2020}.
\begin{figure}[t]
\centering
\includegraphics[width=0.9\linewidth]{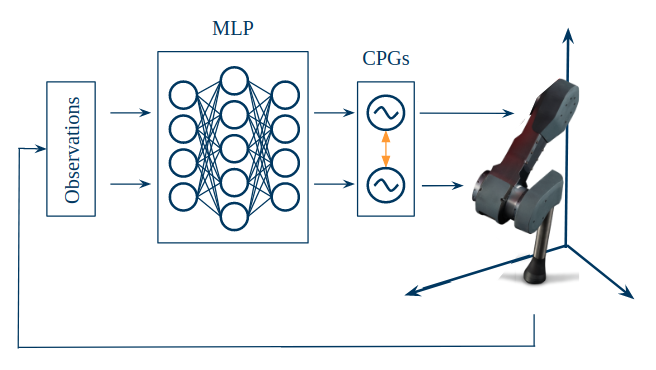}
\caption{The experiments are carried out on a classic \gls{rl} benchmark -- the single-leg hopper -- in a custom environment based on the ANYmal quadruped robot \cite{Hutter2016}. It can hop along the vertical axis and is controlled by \glspl{cpg}. Closed-loop feedback is incorporated using a jointly trained \gls{mlp} network which processes joint sensing observations to reshape the oscillator dynamics of the \glspl{cpg}.
    }
    \label{fig:hopper_architecture}
    \vspace{-15pt}
\end{figure}

These controllers often employ a parametrised policy to map sensory information to low-level actuation commands, and are tuned to optimise a given reward function on data acquired by running the controller itself, which improves with the experience.
In particular, controllers trained using deep-\gls{rl} utilise a \gls{nn} policy to perform this mapping.
As a result, controllers trained with \gls{rl} exhibit behaviours that cannot be hand-crafted by engineers and are further robust to events encountered during the interaction with the environment. 
However, widely-used \gls{nn} architectures, such as \gls{mlp}, do not naturally produce the oscillatory behaviour exhibited in natural locomotion gaits and as such require long training procedures to learn to perform smooth oscillations.


A third family of controllers have been used with promising results for robot locomotion: \glspl{cpg}, a biologically-inspired neural network able to produce rhythmic patterns.
Indeed, the locomotor system of vertebrates is organised such that the \glspl{cpg} -- located in the spine -- are responsible for producing the basic rhythmic patterns, while higher-level centres (the motor cortex, cerebellum, and basal ganglia) are responsible for modulating the resulting patterns according to environmental conditions~\cite{ijspeert2008}.

Besides the intrinsic oscillatory behaviour, several other properties make the usage of \glspl{cpg} desirable for the locomotion task; these include (1) the generation of smooth and rhythmic patterns which are resilient against state perturbations (due to their limit cycle), (2) minimal control dimensionality, i.e.~few high-level signals are needed to control a robot, (3) implementation simplicity (\cref{eqn:cpg_hopf_discr} fully describe the model) and (4) they are model-free, hence well adapted to the locomotion in unknown environments \cite{righetti2008}.
However, very few design principles are available, especially for the integration of sensor feedback in such systems \cite{righetti2008} and, although conceptually promising, we argue that the full potential of \glspl{cpg} has so far been limited by insufficient sensory-feedback integration. 

The ability of Deep-\glspl{nn} to discover and model highly non-linear relationships among the observation -- the inputs -- and control signals -- the outputs -- makes such approaches appealing for control.
In particular, based on Deep-\glspl{nn}, Deep-\gls{rl} demonstrated very convincing results in solving complex locomotion tasks \cite{Hwangbo2019, Joonho2020} and it does not require direct supervision (but rather learns through interaction with the task).  
Hence, we argue that combining Deep-\gls{rl} with \glspl{cpg} could improve the latter's comprehension of the surrounding environment.
However, optimising Deep-\gls{nn} architectures in conjunction with \glspl{cpg} requires adequate methods capable of propagating the gradient from the loss to the parameters, also known as backpropagation. 
In contrast, methodologies that are more commonly applied in tuning \glspl{cpg}, such as \gls{ga}, \gls{pso} and hand-tuning, are rarely used for \gls{nn} applications due to the very high dimensionality of the latter's search space.

Concisely, model-based control requires expert tuning and is computationally demanding during runtime; deep-\gls{rl} controllers are computationally-cheap during runtime, but require offline exploration and ``discovery'' of concepts already known for locomotion (limit cycles, oscillatory behaviour etc.) from scratch, which leads to long training time and careful tuning of reward functions.
\Glspl{cpg}, instead, use concepts developed from bio-inspired sensorimotor control, are computationally cheap during runtime, but are challenging to tune and incorporate feedback within.
To address this, this paper introduces a novel way of using Deep-\glspl{nn} to incorporate feedback into a fully differentiable \gls{cpg} formulation, and apply Deep-\gls{rl} to jointly learn the \gls{cpg} parameters and \gls{mlp} feedback.
%


\subsection{Related Work} \label{sec:related_work}

Our work is related to both the fields of \gls{cpg} design and \gls{rl}, in particular to the application of the latter for the optimisation of the former's parameters.



\glspl{cpg} are very versatile and have been used for different applications including non-contact tasks such as swimmers \cite{ijspeert2007,ijspeert2008}, modular robots \cite{kamimura2003,bonardi2014} and locomotion on small quadrupeds \cite{righetti2008,AjallooeianGay2013,ajallooeian2013,Gay2013}.

The \glspl{cpg} adopted in our research are modelled as Hopf non-linear oscillators (cf. \cref{eqn:cpg_hopf_discr}) which have been successfully transferred to small quadrupedal systems and have exhibited dynamic locomotion behaviours \cite{ajallooeian2013, AjallooeianGay2013, Gay2013}.

The trajectories \glspl{cpg} generate are used as references for each of the actuators during locomotion and a tuning procedure is required to reach coordination.
The optimisation of \gls{cpg}-based controllers usually occurs in simulation through \gls{ga}~\cite{ijspeert2008}, \gls{pso}~\cite{Sprowitz2013, bonardi2014} or expert hand-tuning~\cite{righetti2008,ajallooeian2013,AjallooeianGay2013,Gay2013}.

Prior work has evaluated the performance of \Glspl{cpg} for blind locomotion over flat ground \cite{Sprowitz2013}.
However, to navigate on rough terrain sensory feedback is crucial (e.g. in order to handle early or late contact), as shown in \cite{AjallooeianGay2013}: here, a hierarchical controller has been designed, where \glspl{cpg} relied on a state machine which controlled the activation of the feedback. 
In particular, the stumbling correction and leg extension reflexes are constant impulses triggered by the state machine.
While the attitude control relies on information such as the contact status of each leg, the joint angles read by encoders and the rotation matrix indicating the orientation of the robot's trunk; all these data are processed in a virtual model control fashion and then linearly combined with the \gls{cpg} equations, \cref{eqn:cpg_hopf_discr}. 
Finally, the angle of attack between leg and terrain is useful to accelerate/decelerate the body or locomote on slopes: it is controlled by the sagittal hip joints and it is linearly combined with the equations \cref{eqn:cpg_hopf_discr} to provide feedback.

Similarly to \cite{AjallooeianGay2013}, \cite{Gay2013} also uses feedback, this time based on gyroscope velocities and optical flow from camera to modify the \glspl{cpg} output in order to maintain balance. 
However, in~\cite{Gay2013} the authors first tune \glspl{cpg} in an open-loop setting and then train a \gls{nn} with \gls{pso} to provide feedback (at this stage the parameters of the \glspl{cpg} are kept fixed).
Their method relies on a simple \gls{nn} with 7 inputs -- 4 from the camera/optical flow and 3 from the gyroscope -- and a single hidden layer.
We follow the same design philosophy in the sense that we preprocess the sensory feedback through a \gls{nn}; yet, we propose to tune its parameters in conjunction with the \gls{cpg}.
We argue that in this way the full potential of the interplay of the two can be exploited.
In particular, this effectively allows the feature processing of raw signals to be learnt from experience.


\begin{figure*}[t]
    \centering
    \begin{subfigure}[b]{.32\linewidth}
        \includegraphics[width=\textwidth]{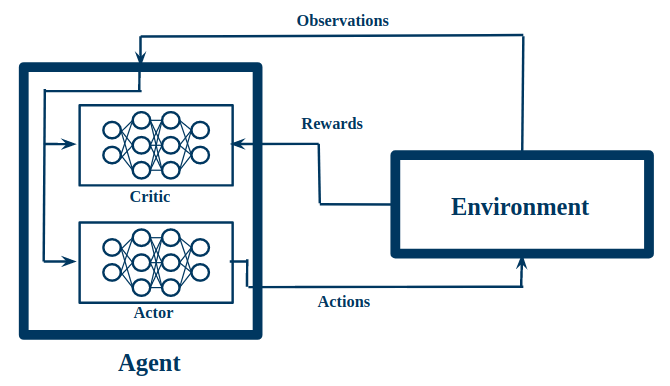}
        \caption{}
        \label{fig:ac_classic}
    \end{subfigure}
    ~
    \begin{subfigure}[b]{.32\linewidth}
        \includegraphics[width=\textwidth]{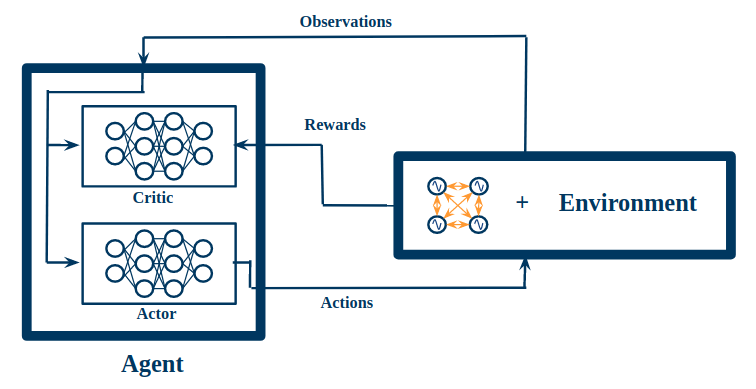}
        \caption{}
        \label{fig:ac_cpg_wrong}
    \end{subfigure}
    ~
    \begin{subfigure}[b]{.32\linewidth}
        \includegraphics[width=\textwidth]{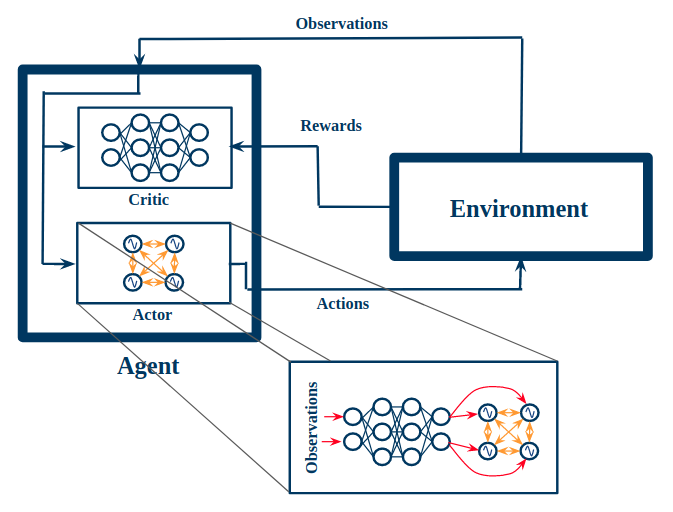}
        \caption{}
        \label{fig:ac_cpg_correct}
    \end{subfigure}
    
    \caption{(\subref{fig:ac_classic}) represents the basic actor-critic Deep-\gls{rl} method adopted for continuous action space control.
    (\subref{fig:ac_cpg_wrong}) illustrates the approach proposed in \cite{Cho2019, Ciancio2011, Nakamura2007, Fukunaga2004}, which consists in a classic actor-critic with \glspl{cpg} embedded in the environment.
    (\subref{fig:ac_cpg_correct}), instead, is the approach proposed in the present work, which includes the \glspl{cpg} alongside the \gls{mlp} network in the actor critic architecture.}
    \label{fig:ac_architectures}
    \vspace{-1em}
\end{figure*}


\Gls{rl} promises to overcome the limitations of model-based approaches by learning effective controllers directly from experience.
%
%
Robotics tasks in \gls{rl} -- such as the hopper considered in this work (\cref{fig:hopper_architecture}) -- are challenging as their action space is continuous and the set of possible actions is infinite. 
Hence, any method based on learning the action values (which are the expected discounted reward received by following a policy) must search through this infinite set in order to select an action.
Differently, actor-critic methods rely on an explicit representation of the policy independent from the value function. 
The policy is known as the actor, because it is used to select actions, while the estimated value function is known as the critic, because it criticises the actions taken by the actor \cite{Sutton1998}, as shown in \cref{fig:ac_classic}.
The critic uses an approximation architecture and simulation to learn a value function, which is then used to update the actor's policy parameters in a direction of performance improvement.
Both of them in Deep-\gls{rl} are classically approximated by \glspl{nn}.


Researchers applied \gls{rl} to optimise \glspl{cpg} in different scenarios \cite{Cho2019, Ciancio2011, Nakamura2007, Fukunaga2004}.
The common factor among them is the formulation of the actor-critic method; yet, they include the \gls{cpg} controller in the environment -- as depicted in \cref{fig:ac_cpg_wrong}. 
In other words, the \gls{cpg} is part of the (black-box) environment dynamics.
According to the authors \cite{Fukunaga2004}, the motivations for including \glspl{cpg} in the environment are their intrinsic recurrent nature and the amount of time necessary to train them, since \glspl{cpg} have been considered \glspl{rnn} (which are computationally expensive and slow to train).
In \cite{Cho2019, Ciancio2011} during training and inference, the policy outputs a new set of parameters for the \glspl{cpg} in response to observations from the environment 
at every time-step.
In this case, the observations processed by the actor network -- which usually represent the feedback -- are responsible for producing a meaningful set of CPG-parameters for the current state.
Conversely, in \cite{Nakamura2007, Fukunaga2004} the parameters are fixed and, similarly to \cite{Gay2013}, \glspl{cpg} receive inputs from the policy.

However, 
whether the \glspl{cpg} parameters were new or fixed every time-step, they all considered \glspl{cpg} as part of the environment rather than making use of their recurrent nature as stateful networks.
We exploit this observation in this paper.



\subsection{Contributions}

In this work, we combine the benefits of \glspl{cpg} and \gls{rl} and present a new methodology for designing \gls{cpg}-based controllers.
In particular, and in contrast to prior work, we embed the \gls{cpg} directly as the actor of an Actor-Critic framework instead of it being part of the environment. 
The advantage of directly embedding a dynamical system is to directly encode knowledge about the characteristics of the task (e.g., periodicity) without resorting to recurrent approaches.
The outcome is \cpgactor, a new architecture that allows end-to-end training of \glspl{cpg} and a \gls{mlp} by means of Deep-\gls{rl}.
In particular, our contributions are:
\begin{compactenum}
    \item For the first time -- to the best of our knowledge -- the parameters of the \glspl{cpg} can be directly trained through state-of-the-art gradient-based optimisation techniques such as \gls{ppo} \cite{Schulman2017}, a powerful \gls{rl} algorithm.
    To make this possible, we propose a fully differentiable \gls{cpg} formulation (\cref{sec:cpg_differentiable_formulation}) along with a novel way for capturing the state of the \gls{cpg} without unrolling its recurrent state  (\cref{sec:how_to_deal_with_recurrent_state}).
    \item Exploiting the fully differentiable approach further enables us to incorporate and jointly tune a \gls{mlp} network in charge of processing feedback in the same pipeline.
    \item We demonstrate a roughly six times better training performance compared with previous state-of-the-art approaches (\cref{sec:results}).
\end{compactenum}

\section{Methodology}\label{methodology}
\begin{figure*}[t]
    \centering
    \begin{subfigure}{\columnwidth}
        \centering
        \includegraphics[width=0.75\textwidth]{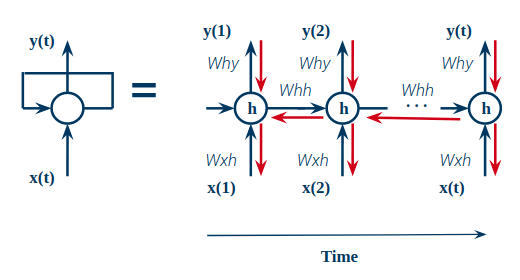}
        \caption{}
        \vspace{5pt}
        \label{fig:backprop_rnn}
    \end{subfigure}
    ~
    \begin{subfigure}{\columnwidth}
        \centering
        \includegraphics[width=0.75\textwidth]{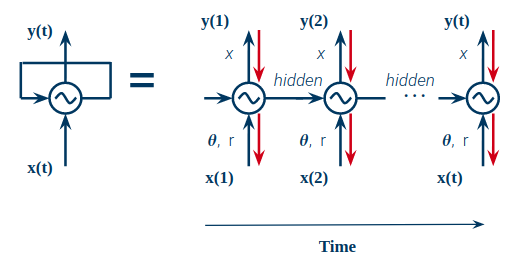}
        \caption{}
        \vspace{5pt}
        \label{fig:backprop_cpg}
    \end{subfigure}
    \caption{The images above show the difference between back-propagation for classic \glspl{rnn} (\ref{fig:backprop_rnn}) and \glspl{cpg} (\ref{fig:backprop_cpg}). In particular to train \glspl{rnn}, the matrices $W_{xh}$, $W_{hy}$, $W_{hh}$ have to be tuned, where $W_{hh}$ regulates the evolution between two \textit{hidden states}. Instead, for \glspl{cpg} only the parameters in $\dot \theta_i$ and $\ddot r_i$ (\cref{eqn:cpg_hopf_discr}) need tuning, while the evolution of the \textit{hidden state} is determined by \cref{eqn:cpg_hidden_state_evolution}.}
    \label{fig:backprop}
    \vspace{-1em}
\end{figure*}

Differently to previous approaches presented in \cref{sec:related_work}, we embed \glspl{cpg} directly as part of the actor in an actor-critic framework as shown in \cref{fig:ac_cpg_correct}.
Indeed, the policy \gls{nn} has been replaced by a combination of an \gls{mlp} network for sensory pre-processing and \glspl{cpg} for action computation, while the value function is still approximated by an \gls{mlp} network.

These measures ensure that the parameters of the \glspl{cpg} are fixed while interacting with the environment and during inference, presenting an alternative (and more direct) way of tuning classical \gls{cpg}-based controllers.

However, a na\"ive integration of \glspl{cpg} into the Actor-Critic formulation is error-prone and special care needs to be taken:
\begin{compactitem}
  \item to attain differentiability through the \gls{cpg} actor in order to exploit gradient-based optimisation techniques;
  \item not to neglect the hidden state as \glspl{cpg} are stateful networks.
\end{compactitem}

We are going to analyse these aspects separately in the following sections.

\subsection{Differentiable Central Pattern Generators}\label{sec:cpg_differentiable_formulation}

Parallel implementations of \gls{rl} algorithms spawn the same policy $\pi_{\theta}$ on parallel instances of the same robot to quickly gather more experiences.
Once the interactions with the simulation environment ends, the observations are fetched in batches and used to update the actor and the critic.
Instead of selecting the best-fitted controller, as \gls{ga} does, the update is based on gradient descent algorithms, such as Adam \cite{Diederik2015}.
Consequently, the implementation of \glspl{cpg} must be differentiable. 

\subsubsection{Hopf Oscillators}
As underlying, oscillatory equation for our \gls{cpg} network, we choose to utilise the Hopf oscillator, as in \cite{ijspeert2007}.
However, since equations in \cite{ijspeert2007} describe a system in continuous time, we need to discretise them for use as a discrete-time robot controller, as in \cref{eqn:cpg_hopf_discr}:
\begin{equation}
\begin{array}{lcl}
    \label{eqn:cpg_hopf_discr}
        \dot \theta_i^t = 2\pi\nu_i(d_i^t) + \zeta_i^t + \xi^t_i
        \\
        \zeta_i^t = \sum_{j} r_j^{t-1} w_{ij} \sin(\theta_j^{t-1} - \theta_i^{t-1} - \phi_{ij})
        \\
        \ddot r_i^t =  a_i(\frac{a_i}{4}(\rho_i(d_i^t) - r_i^{t-1}) - \dot r_i^{t-1}) + \kappa_i^t
        \\
        x_i^t = r_i^t\cos(\theta_i^t)
\end{array}
\end{equation}
where $\cdot^t$ describes the value at the $t$-th time-step, $\theta_i$ and $r_i$ are the scalar state variables representing the phase and the amplitude of oscillator $i$ respectively, $\nu_i$ and $\rho_i$ determine its intrinsic frequency and amplitude as function of the input command signals $d_i$, and $a_i$ is a positive constant governing the amplitude dynamics.
The effects of the couplings between oscillators are accounted in $\zeta_i$ and the specific coupling between $i$ and $j$ are defined by the weights $w_{ij}$ and phase $\phi_{ij}$.
The signal $x_i$ represents the burst produced by the oscillatory centre used as position reference by the motors.
Finally, $\xi_i$ and $\kappa_i$ are the feedback components provided by the \gls{mlp} network.


To calculate the variables $r$ and $\theta$ from their derivative values, we applied a trapezoidal approach, as in \cref{eqn:cpg_hidden_state_evolution}:
\begin{equation}
    \label{eqn:cpg_hidden_state_evolution}
    \begin{array}{lcl}
        \theta^t = \theta^{t-1} + (\dot \theta^{t-1} + \dot \theta^t) \frac{dt}{2}
        \\
        \dot r^t = \dot r^{t-1} + (\ddot r^{t-1} + \ddot r^t) \frac{dt}{2}
        \\
        r^t = r^{t-1} + (\dot r^{t-1} + \dot r^t) \frac{dt}{2}
    \end{array}
\end{equation}
where $dt$ is the timestep duration.

\subsubsection{Tensorial implementation}
The tensorial operations have to be carefully implemented to allow a correct flowing of the gradient and batch computations, both crucial for updating the actor-critic framework.
Let $N$ be the number of \glspl{cpg} in the network, then:
%
\begin{equation}
    \label{eqn:cpg_tensorial_implementation}
    \begin{array}{lcl}
        \dot \Theta^t = 2\pi C_\nu(V, D^t)+ Z^t \mathbf{1} + \Xi^t
        \\
        Z^t = (W V) * (\Lambda R^{t-1}) * \sin(\Lambda \Theta^{t-1} - \Lambda^\intercal \Theta^{t-1} - \Phi V)
        \\
        \ddot R^t =  (A V) * (\frac{A V}{4}(P(V, D^t) - R^{t-1}) - \dot R^{t-1}) + K^t
        \\
        X^t = R^t\cos(\Theta^t)
    \end{array}
\end{equation}

Here, $\Theta \in \mathbb{R}^N$ and $R \in \mathbb{R}^N$ are the vectors containing $\theta_i$ and $r_i$, while $\Xi \in \mathbb{R}^N$ and $K \in \mathbb{R}^N$ contain $\xi_i$ and $\kappa_i$ respectively.
$V \in \mathbb{R}^M$ contains the $M$, constant parameters to be optimised of the network composed by the $N$ \glspl{cpg}.

This said, $C_\nu: \mathbb{R}^M, \mathbb{R}^d \to \mathbb{R}^N$, $P: \mathbb{R}^M, \mathbb{R}^d \to \mathbb{R}^N$ and $A \in \mathbb{R}^{N\times M}$ are mappings from the set $V$ and the command $D^t \in \mathbb{R}^d$ to the parameters that lead $\nu_i$, $\rho_i$ and $a_i$ respectively.

$Z \in \mathbb{R}^{N \times N}$ instead takes into consideration the effects of the couplings of each \gls{cpg} to each \gls{cpg}; all the effect to $i$-th \gls{cpg} will be then the sum of the $i$-th row of $Z$ as in $Z\:\mathbf{1}$, where $\mathbf{1}$ is a vector of $N$ elements with value $1$.
Within $Z$, $W \in \mathbb{R}^{N \times N \times M}$ and $\Phi \in \mathbb{R}^{N \times N \times M}$ extrapolate the coupling weights and phases from $V$, while $\Lambda \in \mathbb{R}^{N \times N \times N}$ encodes the connections among the nodes of the \gls{cpg} network.

\subsection{Recurrent state in CPGs}\label{sec:how_to_deal_with_recurrent_state}
In order to efficiently train \glspl{cpg} in a \gls{rl} setting, we need to overcome the limitations highlighted in \cite{Fukunaga2004}: particularly that \glspl{cpg} are recurrent networks and that \glspl{rnn} take a significant time to train.
In this section, we show how we can reframe \glspl{cpg} as stateless networks and fully determine the state from our observation without the requirement to unroll the \gls{rnn}.

Stateless networks, such as \glspl{mlp}, do not need any information from the previous state to compute the next step and the backpropagation procedure is faster and straightforward.
\Glspl{rnn}, on the other hand, are stateful networks, i.e. the state of the previous time-step is needed to compute the following step output. 
As a consequence, they are computationally more expensive and require a specific procedure to be trained. 
\Glspl{rnn} rely on \gls{bptt}, \cref{fig:backprop_rnn}, which is a gradient-based technique specifically designed to train stateful networks. 
\Gls{bptt} unfolds the \gls{rnn} in time: the unfolded network contains $t$ inputs and outputs, one for each time-step.
As shown in \cref{fig:backprop_rnn}, the mapping from an input $x_t$ to an output $y_t$ depends on three different matrices: $W_{xh}$ determines the transition between the $x_t$ and the hidden state $h$, $W_{hy}$ regulates the transformation from $h_t$ to $y_t$ and lastly $W_{hh}$ governs the evolution between two hidden states.
All the matrices $W_{xh}$, $W_{hy}$, $W_{hh}$ are initially unknown and tuned during the optimisation.
Undeniably, \glspl{cpg} have a recurrent nature and as such require storing the previous hidden state.
However, differently from \glspl{rnn}, the transition between consecutive hidden states in \glspl{cpg} is determined a priori using \cref{eqn:cpg_hidden_state_evolution} without the need of tuning $W_{hh}$.
This observation enables two significant consequences: firstly, \glspl{cpg} do not have to be unrolled to be trained, since, given the previous state and the new input, the output is fully determined.
Secondly, eliminating $W_{hh}$ has the additional benefit of entirely excluding gradient explosion or vanishing during training; both points are illustrated in \cref{fig:backprop_cpg}.
As a result, \glspl{cpg} can be framed as a stateless network on condition that the previous state is passed as an input of the system.


\section{Evaluation} \label{sec:implementation}

\begin{figure*}[ht!]
    \centering

    \begin{subfigure}[b]{.27\linewidth}
        \centering
        \includegraphics[width=\linewidth]{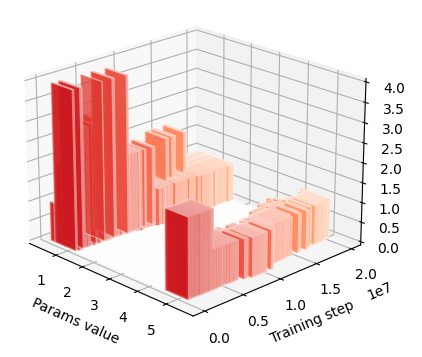}
        \caption{CPG-parameters distribution over time}
        \label{fig:cpg_param_training}
    \end{subfigure}
    ~
    \begin{subfigure}[b]{.29\linewidth}
        \centering
        \includegraphics[width=\linewidth]{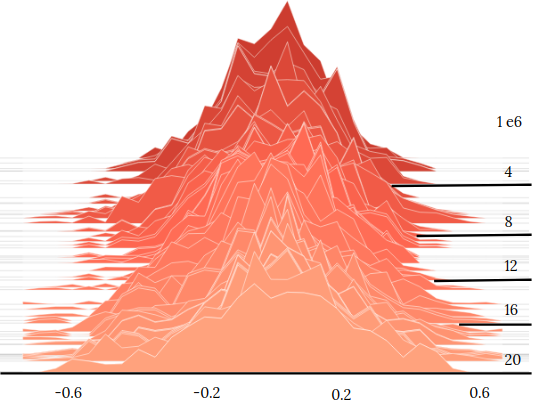}
        \caption{MLP-feedback weights distribution over time}
        \label{fig:mlp_weight_feedback}
    \end{subfigure}
    ~
    \begin{subfigure}[b]{.29\linewidth}
        \centering
        \includegraphics[width=\linewidth]{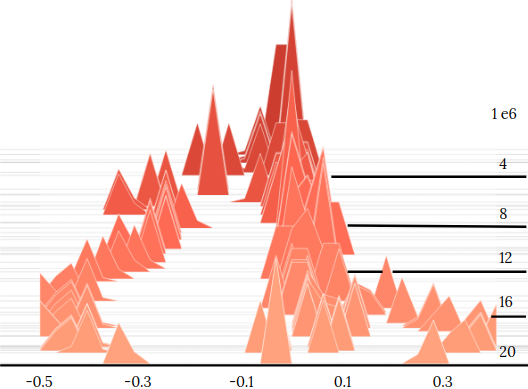}
        \caption{MLP-feedback biases distribution over time}
        \label{fig:mlp_bias_feedback}
    \end{subfigure}
    
    \caption{The set of images above show the evolution -- from darker to lighter colours -- of the distributions of \glspl{cpg} parameters (\cref{fig:cpg_param_training}), weights (\cref{fig:mlp_weight_feedback}) and biases (\cref{fig:mlp_bias_feedback}) of the output layer of \gls{mlp}-feedback network.
    This demonstrates the simultaneous gradient propagation through the \gls{cpg} and \gls{mlp} parameter as described in \cref{sec:cpg_differentiable_formulation}.
    }
    \label{fig:parameters_under_training}
\end{figure*}

The two main components of our approach (\cref{fig:ac_cpg_correct}) are the environment (\cref{fig:hopper_architecture}) and the agent, part of which is \cpgactor.
We evaluate our method on a classic \gls{rl} benchmark: the hopping leg \cite{Fankhauser2013, openaigym2016}, which suits perfectly for \glspl{cpg} as well.
In fact, a single leg needs only two joints to hop and this is the minimal configuration required by coupled Hopf-oscillators to express the complete form; less than two would cancel out the coupling terms, \cref{eqn:cpg_hopf_discr}. 

In order to exclude additional sources of complexity, we tested the efficacy of the new method in the minimal configuration first, however we also plan to address the full locomotion task in the future and developing an environment with realistic masses, forces, inertia and robot's dimensions built a solid base for further development.

Hence, we based the environment on a single leg of the ANYmal quadruped robot, which was fixed to a vertical slider.
Its mass is \SI{3.42}{\kilo\gram} and it is actuated by two series-elastic actuators capable of \SI{40}{\newton\metre} torque and a maximum joint velocity of \SI{15}{\radian\per\second}.
We adopted PyBullet \cite{pybullet} to simulate the dynamics of the assembly and to extract the relevant information.

At every time-step the following observations are captured: the joints' measured positions $p_j^m$ and velocities $v_j^m$, desired positions $p_j^d$, the position $p_h$ and the velocity $v_h$ of the hip attached to the rail.
While the torques $t_j^d$ and the planar velocity of the foot $v_f^{x, y}$ are instead used in computing the rewards, as described in the following lines.
To train \cpgactor, we formulate a reward function as the sum of five distinct terms, each of which focusing on different aspects of the desired system:
\begin{equation}
    \label{eqn:reward}
    \begin{array}{lcl}
        r_1 = (c_1 \cdot \max(v_h, 0))^2
        \\
        r_2 = \sum_{joint}{c_2 \cdot (p^d_j - p^m_j)}^2
        \\
        r_3 = \sum_{joint}{c_3 \cdot (v^m_j)}^2
        \\
        r_4 = \sum_{joint}{c_4 \cdot (t^d_j)}^2
        \\
        r_5 = c_5 \cdot \left\|v_f^{x, y}\right\|
    \end{array}
\end{equation}
where $c_1\ge0$ and $c_2, c_3, c_4, c_5\le0$ are the weights associated with each reward.

In particular, $r_1$ promotes vertical jumping, $r_2$ encourage the reduction of the error between the \textit{desired position} and the \textit{measured position}, $r_3$ and $r_4$ reduce respectively the \textit{measured velocity} and the \textit{desired torque} of the motors and finally, $r_5$ discourage the foot from slipping.

Although the \cpgactor has been extensively treated in \cref{methodology}, it is important to strengthen that it has been integrated in an existing \gls{rl} framework based on \textit{OpenAI Baselines} \cite{baselines}.
This allows to exploit standard, well-tested \gls{rl} implementations, parallel environments optimisation, GPU-computations and allows to extend the approach to other algorithms easily as they share the same infrastructure.

\subsection{Experimental setup}
\cpgactor is compared against \cite{Cho2019} using the same environment.
Both the approaches resort to an actor-critic formulation, precisely running the same critic network with two hidden layers of 64 units each.
Indeed, the main difference is the actor, which is described in detail in \cref{methodology} for the \cpgactor case, while \cite{Cho2019} relies on a network with two hidden layers of 64 units each.
%

As \cref{sec:results} illustrates, an appropriate comparison between \cpgactor and \cite{Cho2019} required the latter to be warm-started to generate \textit{desired positions} resulting in visible motions of the leg.
Differently from the salamander \cite{ijspeert2007}, already tuned parameters are not available for the hopping task, hence a meaningful set from \cite{AjallooeianGay2013} was used as reference.
The warm-starting consisted in training the actor network for 100 epochs in a supervised fashion using as target the aforementioned parameters.

\begin{figure*}[ht]
    \centering
    \begin{subfigure}{.44\linewidth}
        \includegraphics[width=\textwidth]{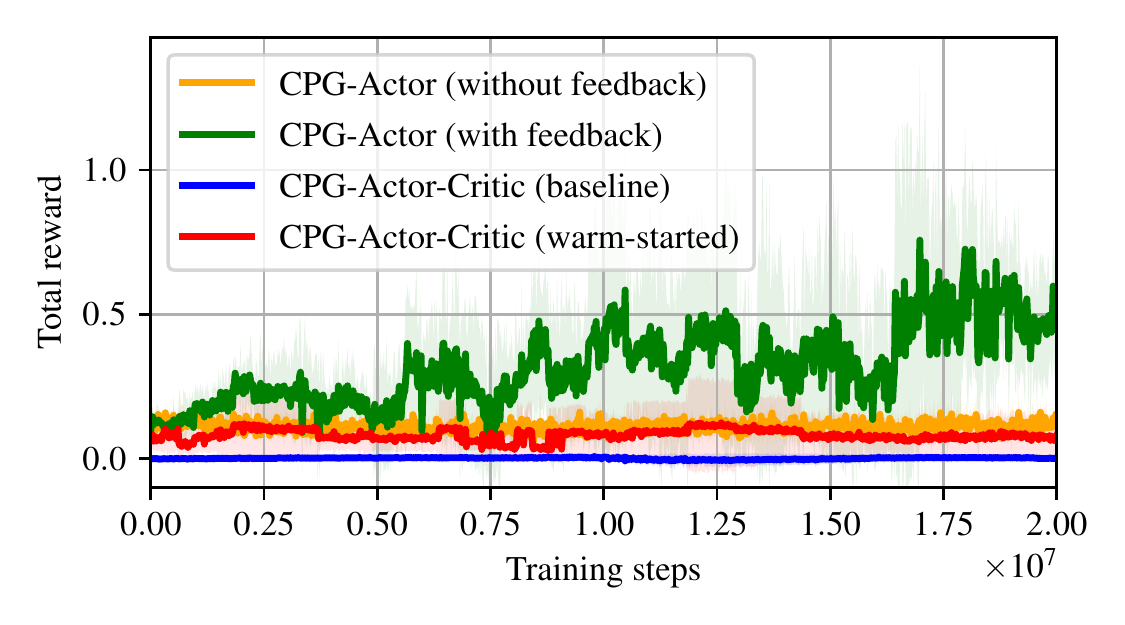}
        \caption{Episode reward over 20M time steps horizon.}
        \label{fig:ep_reward}
    \end{subfigure}
    ~
    \begin{subfigure}{.44\linewidth}
        \includegraphics[width=\textwidth]{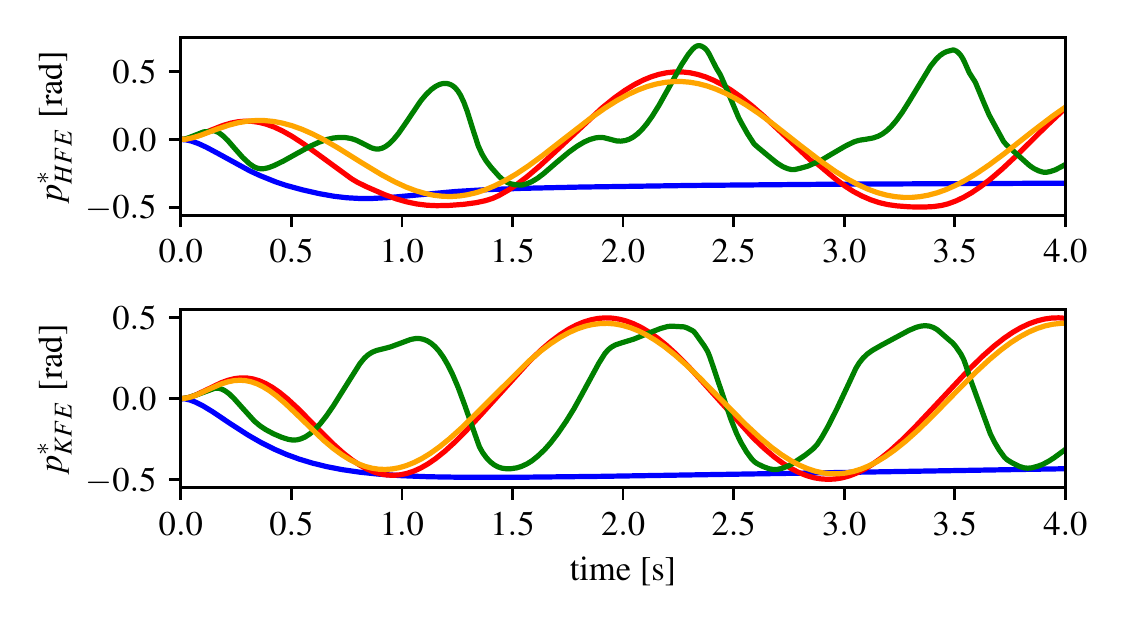}
        \caption{Desired positions generated by \gls{cpg}-Actor-Critic \cite{Cho2019} and \cpgactor.}
        \label{fig:cpg_output}
    \end{subfigure}
    ~
    \begin{subfigure}{.44\linewidth}
        \includegraphics[width=\textwidth]{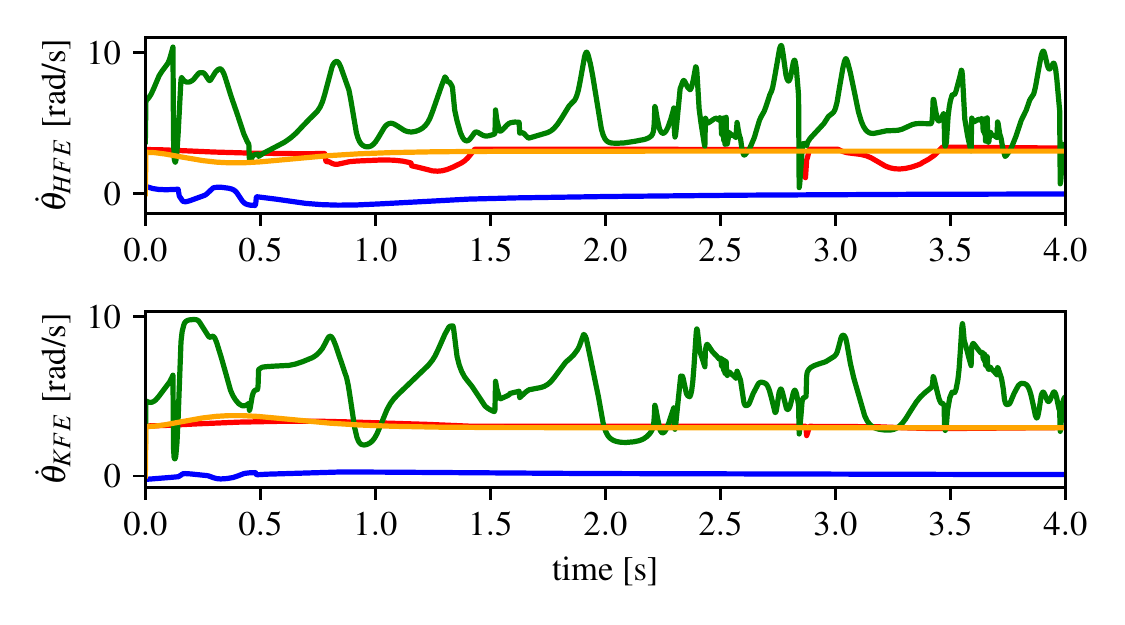}
        \caption{Comparison between $\dot \theta$, \cref{eqn:cpg_hopf_discr}, generated by \gls{cpg}-Actor-Critic \cite{Cho2019} and \cpgactor.}
        \label{fig:cpg_theta_dot}
    \end{subfigure}
    ~
    \begin{subfigure}{.44\linewidth}
        \includegraphics[width=\textwidth]{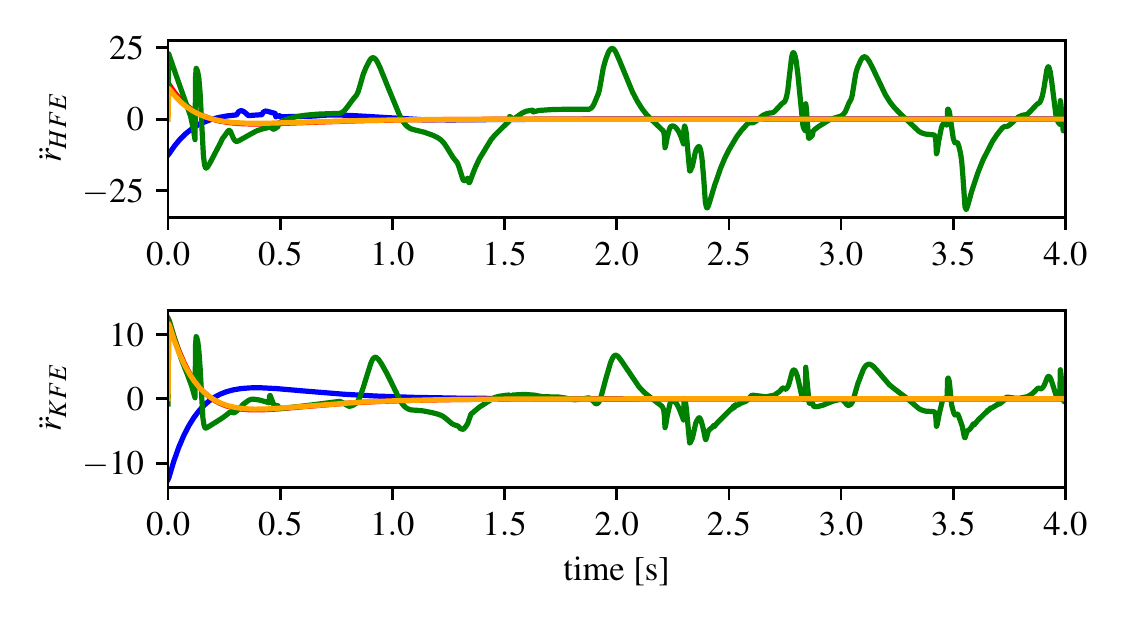}
        \caption{Comparison between $\ddot r$, \cref{eqn:cpg_hopf_discr}, generated by \gls{cpg}-Actor-Critic \cite{Cho2019} and \cpgactor.}
        \label{fig:cpg_r_dot_dot}
    \end{subfigure}
    ~
    \caption{(\ref{fig:ep_reward}) represents how the reward evolves during training, each of the approaches has been run for five times and the rewards averaged.
    (\ref{fig:cpg_output}) illustrates the trajectories generated by the different approaches: \cite{Cho2019} with warm-start produces an output similar to \cpgactor without feedback. While \cpgactor with feedback presents a heavily reshaped signal.  
    The different contribution of the feedback in the two aforementioned approaches is explained by (\ref{fig:cpg_theta_dot}) and (\ref{fig:cpg_r_dot_dot}), which are the phase and amplitude equations in \cref{eqn:cpg_hopf_discr}. Here the feedback -- in \cpgactor case -- is actively interacting with the controller according to the state observed, resulting into visibly reshaped $\dot \theta$ and $\ddot r$ (green lines).}
    \label{fig:results}
    \vspace{-1em}
\end{figure*}

\begin{figure*}[ht]
    \centering
    \begin{subfigure}{0.49\linewidth}
        \centering
        \includegraphics[width=\linewidth]{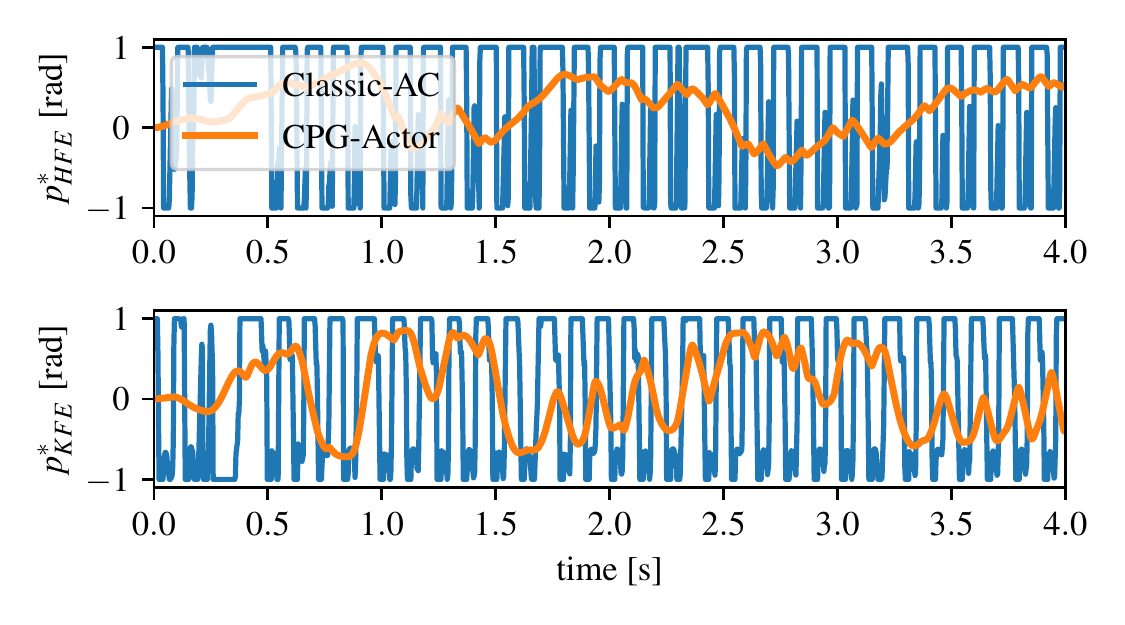}
        \caption{}
        \label{fig:des_pos}
    \end{subfigure}
    \begin{subfigure}{0.49\linewidth}
        \centering
        \includegraphics[width=\linewidth]{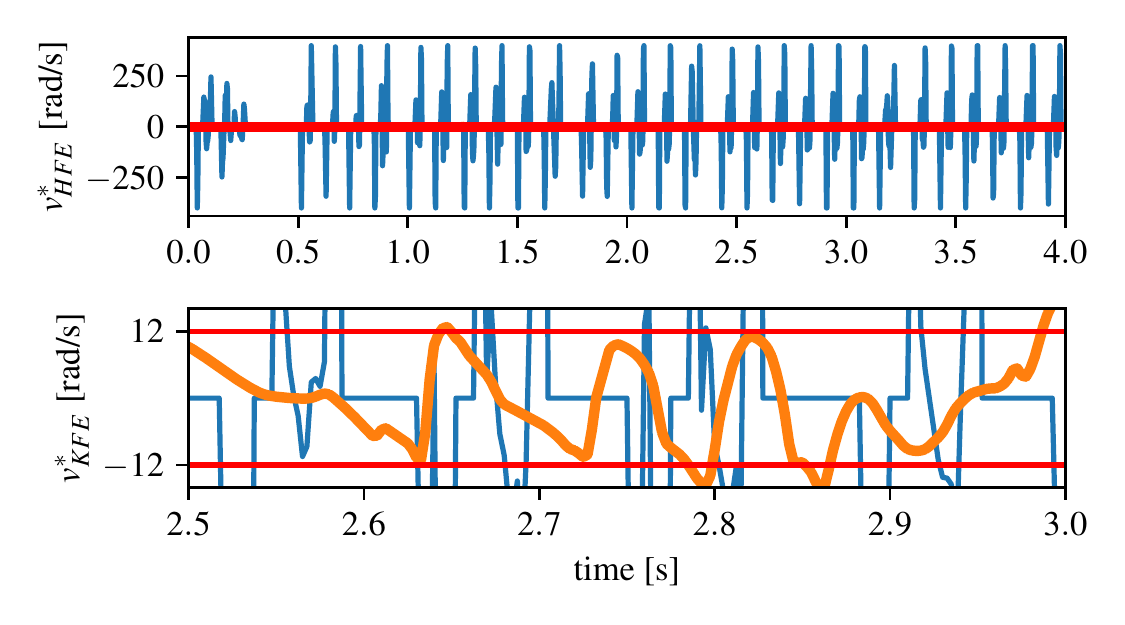}
        \caption{}
        \label{fig:des_vel}
    \end{subfigure}
    
    \caption{The images above compare the desired position (\ref{fig:des_pos}) and the desired velocity (\ref{fig:des_vel}) generated by \glspl{cpg} and \gls{mlp}. The plot relative to the knee joint (KFE) (\ref{fig:des_vel}) is magnified to better show the sharp output of the \gls{mlp} and how \gls{cpg}'s desired velocities are very close to the motors' limits (horizontal red lines), even if the latter were not explicit constraints of the optimisation.}
    \label{fig:des_output}
    \vspace{-1em}
\end{figure*}


\section{Results} \label{sec:results}
\subsection{Validation of end-to-end training}
We first demonstrate the effectiveness of \cpgactor for end-to-end training.
\Cref{fig:parameters_under_training} shows how the parameters belonging to both the \gls{cpg} controller (\cref{fig:cpg_param_training}) and the network that processes the feedback (\cref{fig:mlp_weight_feedback,fig:mlp_bias_feedback}) evolve in conjunction.
This is signified by their distributions changing over the course of the learning process, from darker to lighter shades as the training process proceeds. 

\subsection{Comparison between MLP and \cpgactor}
In \cref{fig:des_output}, the \textit{desired positions} and the \textit{desired velocities} of a classic actor-critic (as in \cref{fig:ac_classic}) and of the \cpgactor are compared after training for the same amount of time.
What emerges is that the \textit{desired positions} of the \cpgactor is smooth (\cref{fig:des_pos}), while the \gls{mlp}-actor shows a bang-bang behaviour.
Moreover, the \textit{desired velocities} (\cref{fig:des_vel}) of the \cpgactor are almost respecting the motor's operative range -- red horizontal lines -- without explicitly constraining the optimisation around these values.
The \textit{desired positions} and \textit{desired velocities} generated by \cpgactor -- under the same setup used for the \gls{mlp} -- appear to be closer to a safe deployment on a real robot compared to a classic actor-critic.
Despite a more careful tuning of the rewards could produce desirable trajectories for the \gls{mlp} as well, \glspl{cpg} require less efforts to achieve a smooth oscillatory motion and this is the reason behind investigating their potential.

\subsection{\cpgactor and previous baselines, comparison}
Since the integration of \glspl{cpg} in \gls{rl} frameworks has already been proposed in other works, we validated our approach against \cite{Cho2019} to prove its novelty and the resulting advantages.
The approach proposed in \cite{Cho2019} is applied to a salamander robot, to better adapt the original version, presented in \cite{ijspeert2007}, to more challenging environments. 
Hence, the integration of exteroceptive information to improve the capabilities of the controller is pivotal.

We reproduced the aforementioned method and applied it to our test-bed, in \cref{fig:hopper_architecture}, to compare it with \cpgactor.
Warm-starting the policy network referring to the parameters in \cite{ijspeert2007} is one of the steps proposed in \cite{Cho2019} and the result on our hopping leg is represented by the red line (\cref{fig:ep_reward}).
The warm-starting is a crucial operation, because, without it, the outcome (blue line, \cref{fig:ep_reward}) would have not been adequate for a fair comparison with CPG-Actor, due to its poor performances.
Conversely, Cpg-Actor (green line, \cref{fig:ep_reward}) functions in average better along training than the other approaches, reaching roughly six time more reward after 20 million time-steps.

We investigated the reason of such different performances and we argue it lies in the way the feedback affects the \gls{cpg} controller.
\Cref{fig:cpg_theta_dot,fig:cpg_r_dot_dot} represent the evolution over time of the \glspl{cpg} (\cref{eqn:cpg_hopf_discr}).
Observing $\dot \theta$ and $\ddot r$ in experiments with \cite{Cho2019} it is evident they do not show responsiveness to the environment, since the blue and the red lines remain almost flat during the whole episode.
On the other hand, $\dot \theta$ and $\ddot r$ in \cpgactor experiments (green line) demonstrate substantial and roughly periodic modifications over time.
This is also suggested by the \textit{desired positions} in  \cref{fig:cpg_output}: in the case of \cpgactor the original \gls{cpg}'s cosine output is heavily reshaped by the feedback, while \cite{Cho2019} presents almost a sinusoidal behaviour.

Besides, we compared our approach without feedback (orange line) with \cite{Cho2019} and it surprisingly performs better than the latter.
This is quite impressive since \cite{Cho2019} updates its output based on the observations received, while \cpgactor was tested in open-loop.

\subsection{Evaluation of progressive task achievement}
\begin{figure*}[ht!]
    \centering
    \begin{subfigure}[b]{.32\linewidth}
        \centering
        \includegraphics[width=\linewidth]{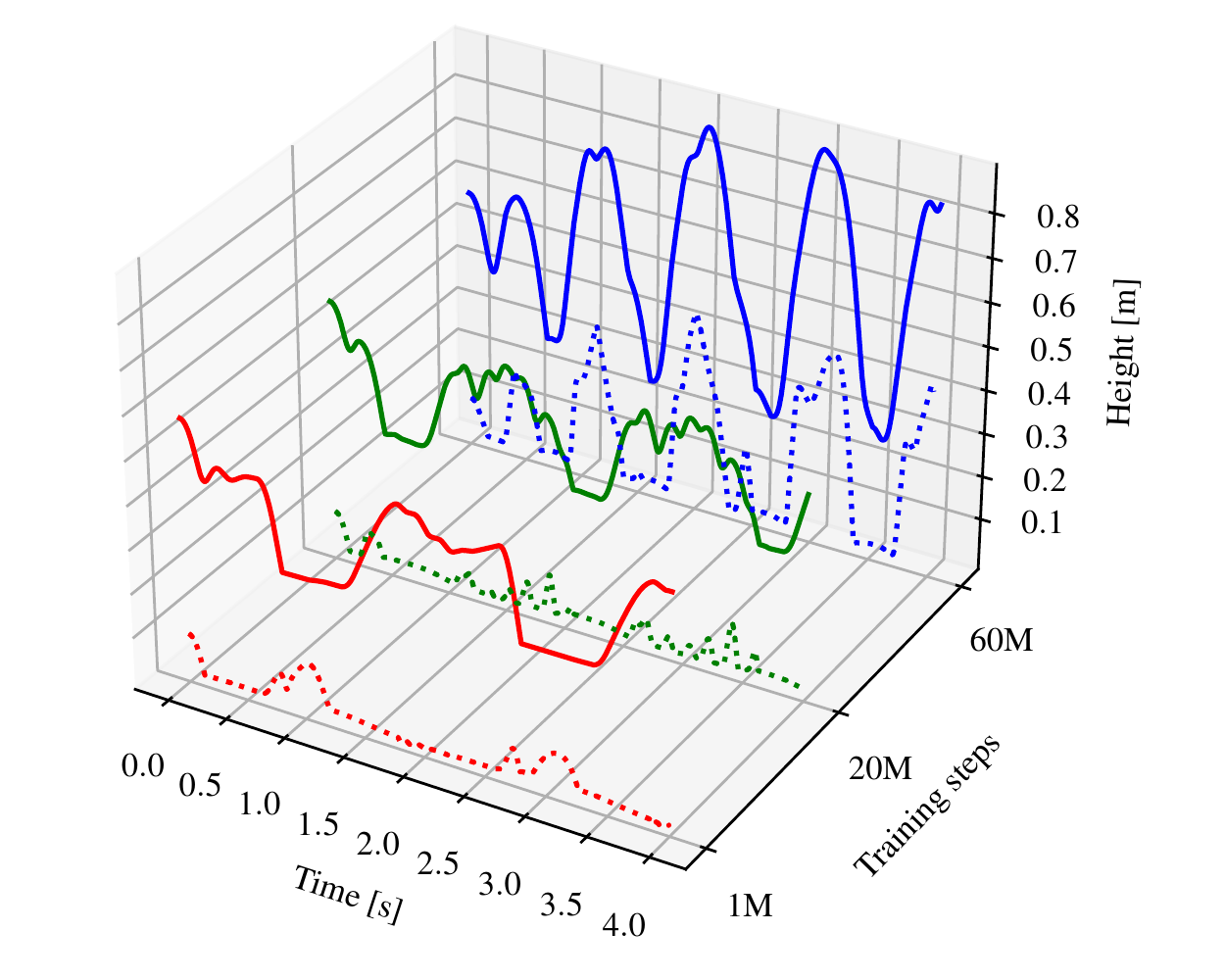}
        \caption{The hopper progressively learns to jump.}
        \label{fig:hopper_hip_foot_pos}
    \end{subfigure}
    ~
    \begin{subfigure}[b]{.32\linewidth}
        \centering
        \includegraphics[width=\linewidth]{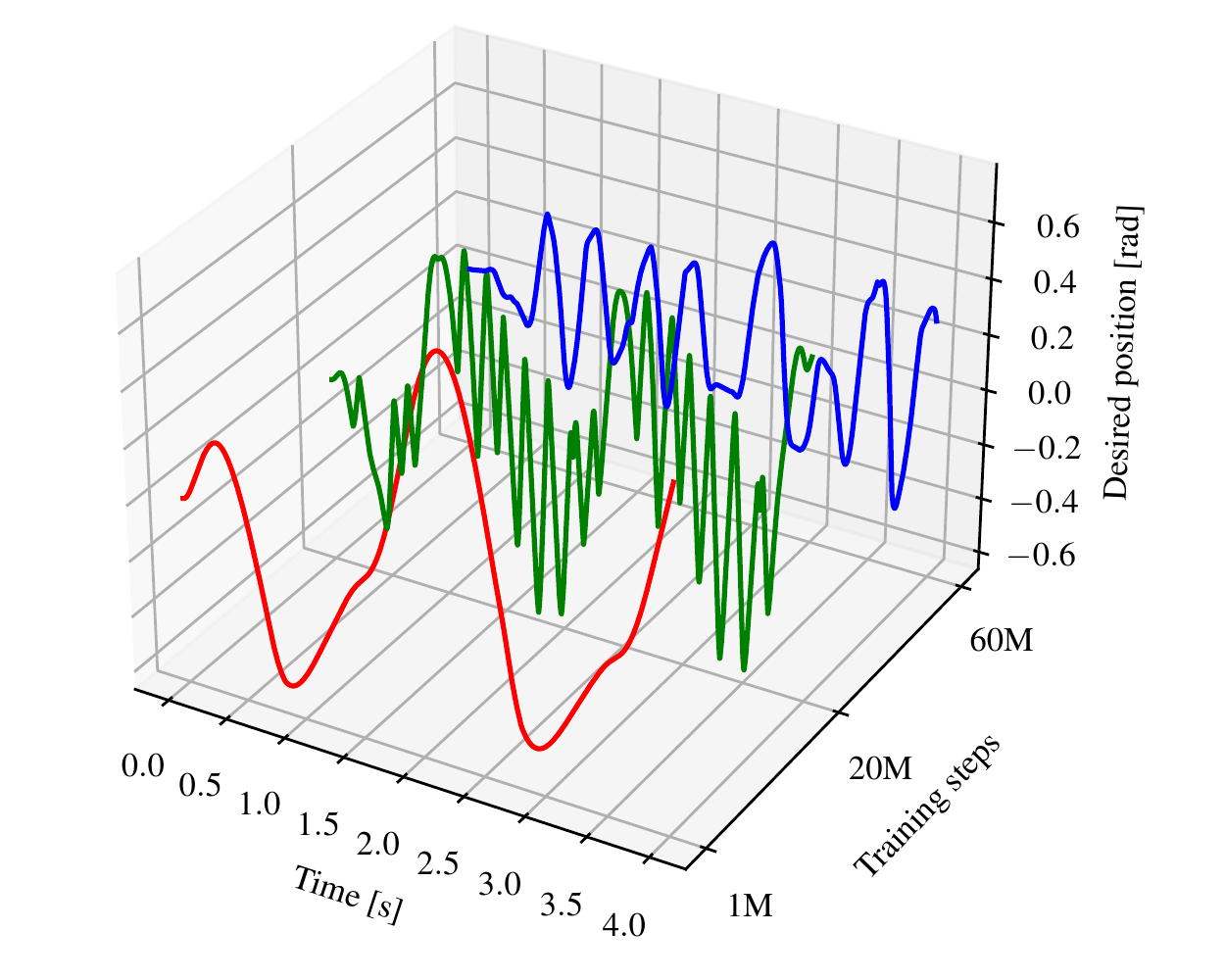}
        \caption{Desired hip positions across epochs}
        \label{fig:hfe_during_training}
    \end{subfigure}
    ~
    \begin{subfigure}[b]{.32\linewidth}
        \centering
        \includegraphics[width=\linewidth]{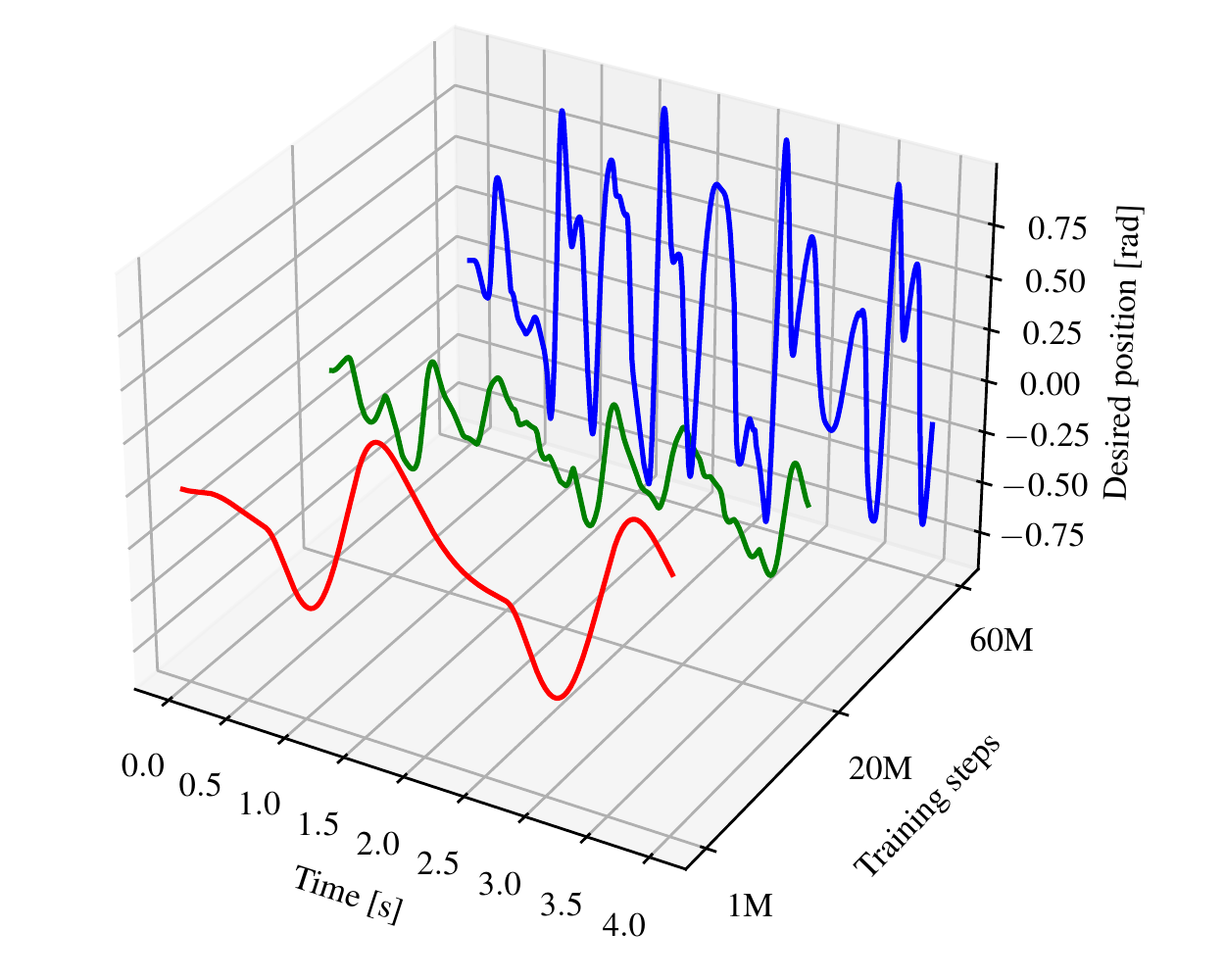}
        \caption{Desired knee positions across epochs}
        \label{fig:kfe_during_training}
    \end{subfigure}
    \caption{
        The above figures demonstrate that the CPG-Actor progressively learns to jump indicated by higher peaks of both the hip (solid line) and foot (dotted line) heights (\cref{fig:hopper_hip_foot_pos}).
        We further show the evolution of the output of the oscillators across epochs (\cref{fig:hfe_during_training,fig:kfe_during_training}).
    }
    \label{fig:cpg_actor_training_over_time}
\end{figure*}

The last set of experiments presented assess how \glspl{cpg}' outputs and the overall behaviour evolve over the course of the learning.
The plots in \cref{fig:cpg_actor_training_over_time} present 
the system at \num{1}, \num{20} and \num{60} million time-steps of training.
In particular, \cref{fig:hfe_during_training,fig:kfe_during_training} are very similar, since they represent how the position output of reciprocally hip (HFE) and knee (KFE) joints develop over time.
\Cref{fig:hopper_hip_foot_pos}, instead, shows the progress of the hopper in learning to jump; indeed, the continuous and dotted lines -- respectively indicating the hip and the foot position -- start quite low at the beginning of the training, to almost double the height after \num{60} millions time-steps.


\section{Discussion and Future work} \label{sec:conclusions}
We propose \cpgactor, an effective and novel method to tune \gls{cpg} controllers through gradient-based optimisation in a \gls{rl} setting.

In this context, we showed how \glspl{cpg} can directly be integrated as the Actor in an Actor-Critic formulation and additionally, we demonstrated how this method permits us to include highly non-linear feedback to reshape the oscillators' dynamics.

Our results on a locomotion task using a single-leg hopper demonstrated that explicitly using the \gls{cpg} as an Actor rather than as part of the environment results in a significant increase in the reward gained over time compared with previous approaches.

Finally,  we  demonstrated  how  our  closed-loop  \gls{cpg}  progressively  improves  the  hopping  behaviour relying only on basic reward functions.

In the future, we will extend the present approach to the full locomotion task and deploy it on real hardware.
In fact, we believe this novel approach gives \glspl{cpg} all the tools to rival state-of-the-art techniques in the field and gives researchers a less reward-sensitive training method.

%

\section*{Acknowledgements}
The authors would like to thank Prof. Auke Ijspeert and his students, Jonathan Arreguit and Shravan Tata, for 
providing insights and feedback.
We further would like to thank Alexander Mitchell for his feedback in reviewing the manuscript.

\bibliographystyle{IEEEtran}
\bibliography{IEEEabbrv,IEEEconf,bibliography}

\end{document}